\documentclass[letterpaper]{article}
\usepackage{spconf,amsmath,graphicx}

\title{Automated Segmentation of Cervical Nuclei in Pap Smear Images Using Deformable Multi-path Ensemble Model}
\name{Jie Zhao$^{1, 2}$, Quanzheng Li$^{1, 2, 3, 4}$, Xiang Li$^{3, 4}$, Hongfeng Li$^{1, 2}$,  Li Zhang$^{\star, 1, 2}$\thanks{$^{\star}$Corresponding author: Li Zhang, zhangli\underline{ }pku@pku.edu.cn.}}
\address{$^{1}$ Center for Data Science in Health and Medicine, Peking University, Beijing, China; 
	\\$^{2}$ Beijing Institute of Big Data Research, Beijing, China; 
	\\$^{3}$ MGH/BWH Center for Clinical Data Science, Boston, MA 02115, USA;
    \\$^{4}$ Department of Radiology, Massachusetts General Hospital, Boston, MA 02115, USA;}
\begin{document}
%\ninept
\maketitle

\begin{abstract}
Pap smear testing has been widely used for detecting cervical cancers based on the morphological properties of cell nuclei in microscopic image. An accurate nuclei segmentation could thus improve the success rate of cervical cancer screening. In this work, a method of automated cervical nuclei segmentation using Deformable Multipath Ensemble Model (D-MEM) is proposed. The approach adopts a U-shaped convolutional network as a backbone network, in which dense blocks are used to transfer feature information more effectively. To increase the flexibility of the model, we then use deformable convolution to deal with different nuclei irregular shapes and sizes. To reduce the predictive bias, we further construct multiple networks with different settings, which form an ensemble model. The proposed segmentation framework has achieved state-of-the-art accuracy on Herlev dataset with Zijdenbos similarity index (ZSI) of 0.933$\pm$0.14, and has the potential to be extended for solving other medical image segmentation tasks.
\end{abstract}


\section{Introduction}
\label{sec:intro}
Pap smear test is extensively used in gynecology to screen premalignant and malignant diseases in the cervix. The 5 typical abnormal results \cite{insinga2004diagnoses} of this test include Atypical squamous cells of undetermined significance (ASC-US), Low-grade squamous intra-epithelial lesion (LSIL), High-grade squamous intra-epithelial lesion (HSIL), Atypical squamous cells, cannot exclude HSIL (ASC-H), and Atypical glandular cells (AGC). All these results point to symptoms accompanying with different levels of nuclear disorders, which contain substantial diagnostic information for cervical diseases. In order to accurately analyze this information for  efficient cervical disease screening, an accurate segmentation of nuclei is necessary.

\begin{figure*}[!htp]
	\centering
	\centerline{\includegraphics[width=0.95\linewidth]{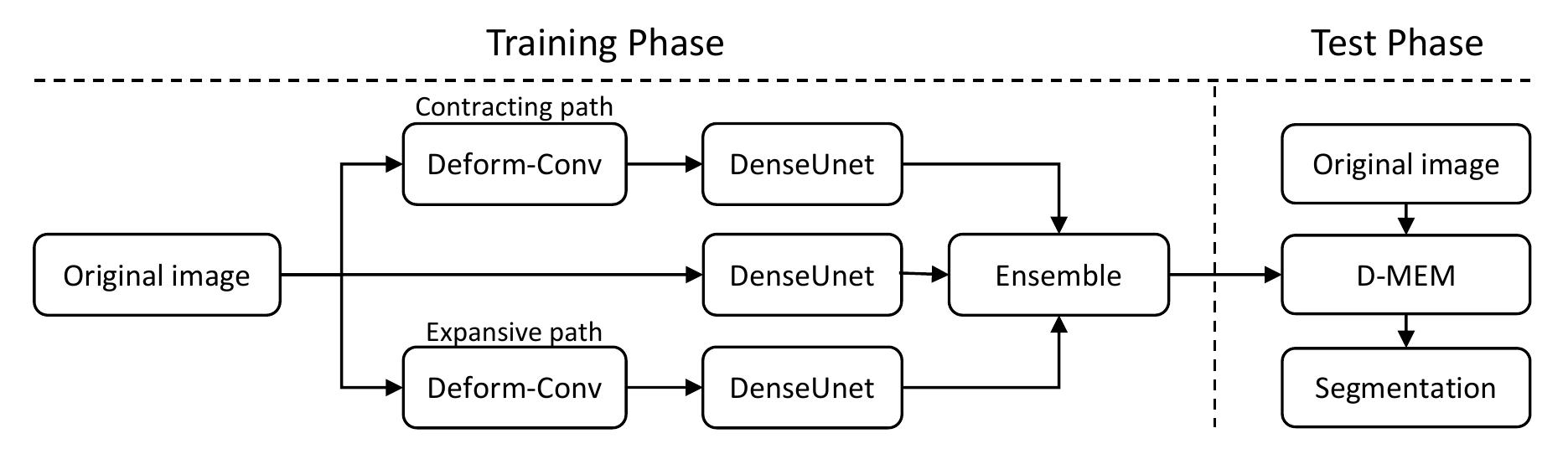}}
	%  \vspace{2.0cm}
	\caption{Flowchart of the experimental procedure.}
	\label{fig:fig_flowchart}
\end{figure*}

Methods for segmenting cervical nuclei include two major types: image-understanding approaches and deep-learning based approaches. Examples of image-understanding approaches are K-means clustering \cite{li2012cytoplasm}, watersheds \cite{huang2010effective}, adaptive thresholding \cite{plissiti2011automated}, active contour model (snake) \cite{bamford1998unsupervised}, morphological operations \cite{wang2005computerized} and graph-cuts \cite{zhang2014segmentation}. An obvious drawback of these methods is their insufficiency to fully describe cervical nuclei, as the methods are often based on an incomplete hand-crafted set of low-level features. In addition, the low-level features themselves lack detailed structural information and lead to poor performance in segmentation.  Thus, to guarantee the quality of segmentation, multiple methods are required for different types of cervical nuclei,  and so do several pre- and post-processing. However, the long pipe-lines and the complex process flow suffer from instability. Any segmentation errors in the intermediate steps may result in a failure of the entire segmentation process.

On the other hand, several research groups have attempted to perform cervical nuclei segmentation using deep learning frameworks, as encouraged by the recent vast success of deep learning approaches across many different computer vision tasks. Song Y et al. present a supervised deep learning network based on super-pixel strategy for segmenting cervical cytoplasm and nuclei \cite{song2014deep}.  The same group also reports a nuclei segmentation by combining a multiscale convolutional network (MSCN) and graph-partitioning approach \cite{song2015accurate}. Sharma B. et al. present a way of using fuzzy c-means (FCM) clustering and back propagation neural network (BPNN) for segmenting cervical images, where the FCM clustering results serve as extra features of BPNN \cite{sharma2016improved}. Zhang et al. propose a two-stage segmentation approach that combines fully convolutional networks (FCN) and dynamic programming (called FCN-G), in which  FCN results are treated as initials of graph-based segmentation \cite{zhang2017combining}. These methods substantially improve segmentation accuracies as opposed to traditional image-understanding approaches, but still suffer from pipelined process and usually have task-specific or data-specific network structures. 

\begin{figure}[!t]
	\centering
	\centerline{\includegraphics[width=0.95\linewidth]{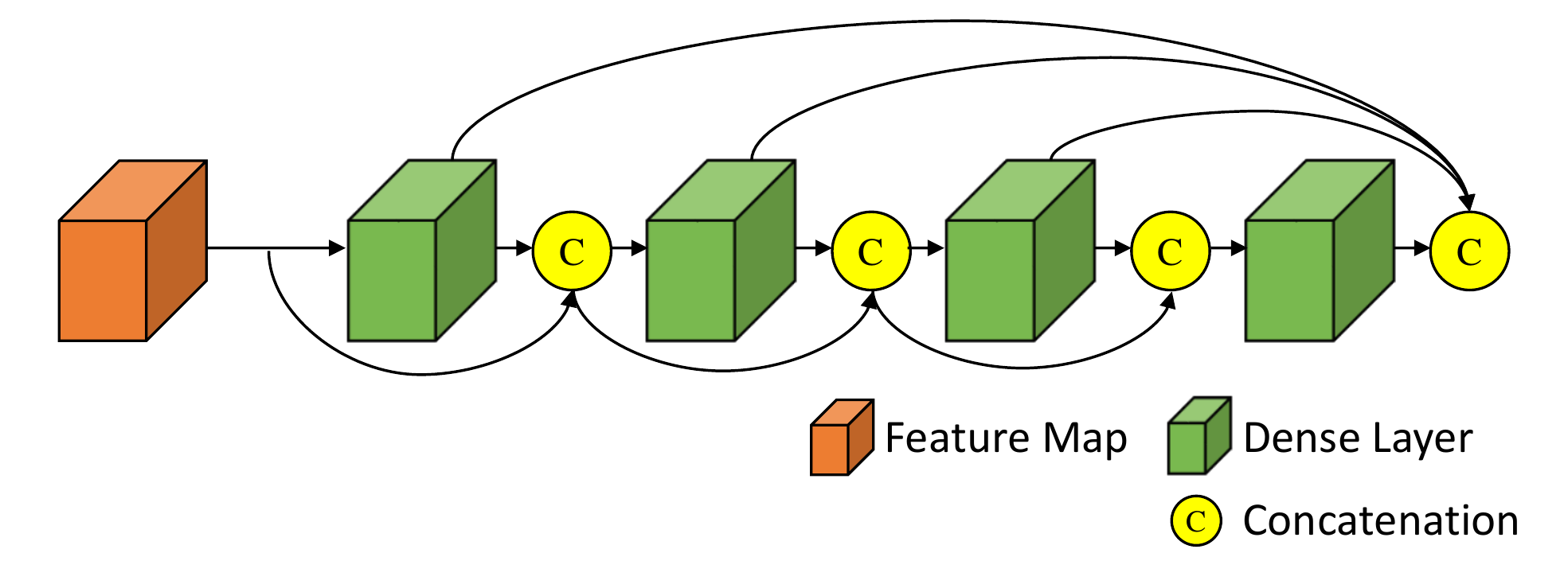}}
	%  \vspace{2.0cm}
	\caption{Illustration of the architecture of a Dense block.}
	\label{fig:Dense Block}
\end{figure}

In order to solve the aforementioned problems, we propose segmenting cervical nuclei via Deformable Multipath Ensemble Model (D-MEM) based on novel deep neural networks. Building a U-shaped network (Unet), we adopts dense block structure with deformable convolutions to improve the ability of recognizing detailed structures in cytological images. A parallel ensemble strategy is introduced in order to reduce the predictive bias from training a single network. The model is also trained end-to-end without pre- and post-processing, which avoids the issues intrigued by pipelined structures. The proposed D-MEM is evaluated on the Herlev dataset\cite{jantzen2005pap} and shows superior performance. 
\vspace*{-0.09cm}
\section{Methods}
\label{sec:Proposed Method}
Three important model structures are introduced to improve the accuracy in the proposed D-MEM: 1) Dense blocks are exploited, which improve the information flow between neural network layers and show better capability of feature extraction, and also feature reuse. 2) Since the abnormal cervical nuclei may display irregular shape rather than circular shape of normal nuclei, the model needs to be more sensitive to subtle changes of the objects. We thus add deformable convolutions to capture detailed structures of nuclei. 3) A single network may suffer from predictive bias and result in coarse segmentation during inference. To further improve the segmentation process, we organize the model in a multi-path fashion, which trains multiple networks simultaneously with different settings and integrates the results using a majority voting strategy. Fig \ref{fig:fig_flowchart} shows the flow chart of this method comprises two major parts: contracting path and expansive path.

\subsection{Dense Blocks}
Traditional convolutional layers form a sequential feed-forward network structure, in which the output of the $l$th layer is taken as the input of the $(l+1)$th layer. The transition can be represented as $X_{l+1}=H_{l+1}(X_l)$, where $X_l$ and $X_{(l+1)}$ denote inputs of $l$th layer and $(l+1)$th layer, respectively; $H_{l+1}$ denotes the mapping.

Under this sort of network architecture, information is transferred by feed-forward pass and model weights are updated by back propagation. However, the architecture suffers from occasional gradient vanishing issues. ResNets attempt to solve this issue with skip-connections that bypass the convolution of feature maps (with non-linear activation) with an identity transition, $X_{l+1}=H_{l+1}(X_l)+X_l$.

The skip-connection allows the gradient directly flows through different layers, which maintains the information magnitude (avoiding information vanishing during training process). To further enhance the connection between layers, dense blocks are introduced. As shown in Fig \ref{fig:Dense Block}, the $l$th layer receives the feature maps of all preceding layers in the same dense block, 
\begin{equation}
X_l=H_l([X_{l-1},...,X_1,X_0]) 
\end{equation}
where $[X_{l-1},...,X_1,X_0]$ represents the concatenation of the feature maps from the preceding layers. In this case, information transition is more compact.

\begin{figure}[!htb]
	\centering
	\centerline{\includegraphics[width=0.95\linewidth]{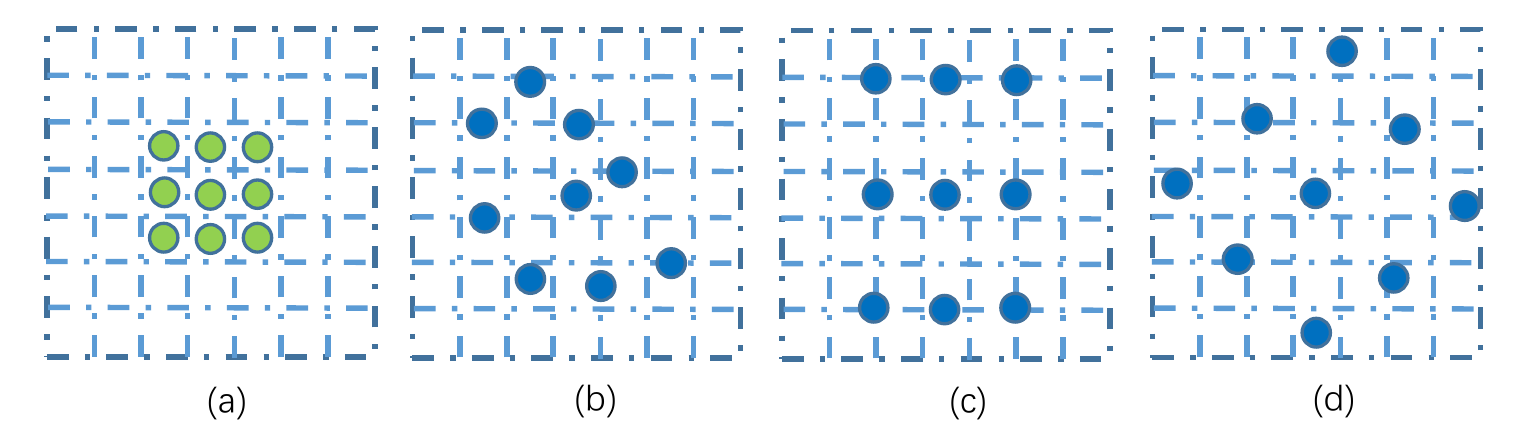}}
	%  \vspace{2.0cm}
	\caption{Illustration of the deformable convolution with a kernel size of 3. (a) normal convolution; (b) deformable convolution with arbitary offsets; (c-d) special case of deformable convolution.}
	\label{fig:deform-conv}
\end{figure}

\subsection{Deformable Convolutional Layers}
An essential task of our proposed model is to segment cervical nuclei with irregular shapes and different sizes. However, the traditional Unet is inherently limited for dealing with object transformation due to its regular convolutional kernel of fixed shape. The deformable convolution is thus introduced to enhance the transformation modeling capability for adjusting the receptive fields of the convolutional kernels \cite{jegou2017one}. 

In classic CNN architecture, as shown in the Fig \ref{fig:deform-conv}(a), the regular convolutional grid $R$ of 3$\times$3 kernel is defined as $R=\left\{(-1,-1),(-1,0),...,(0,1),(1,1)\right\}$.
The value of output feature map ($y$) at position $P_0$ is strictly computed as a weighted summation depending on the grid:
\begin{equation}
y(P_0)=\sum_{P_n\in{R}}w(P_n)\cdot{x(P_0+P_n)}
\end{equation}
In the deformable convolution, small offsets $\Delta P_n$ ($n = 1,2,\dots,N$) are introduced for adjusting the spatial locations of the convolutional inputs (see Fig \ref{fig:deform-conv}(b-d)). The value of output feature map at position $P_0$ could then be defined as:
\begin{equation}
y(P_0)=\sum_{P_n\in{R}}w(P_n)\cdot{x(P_m)}
\end{equation}
where $P_m$ represents a fractional position $(P_m = P_0+P_n+\Delta P_n)$ and $x(P_m)$ is computed using bilinear interpolation:
\begin{equation}
x(P_m)=\sum_{q}G(q,P_m)\cdot{x(q)}
\end{equation}
where $q$ enumerates all integral spatial locations in the neighborhood of $P_m$, and $G(\cdot)$ is the bilinear interpolation kernel.
Similar to the attention mechanics, the deformable convolution networks are able to optimize the offsets for subtle structure by augmenting the spatial sampling locations on feature maps \cite{ouyang2015deepid}. 

\begin{figure}[!htb]
	\centering
	\centerline{\includegraphics[width=0.6\linewidth]{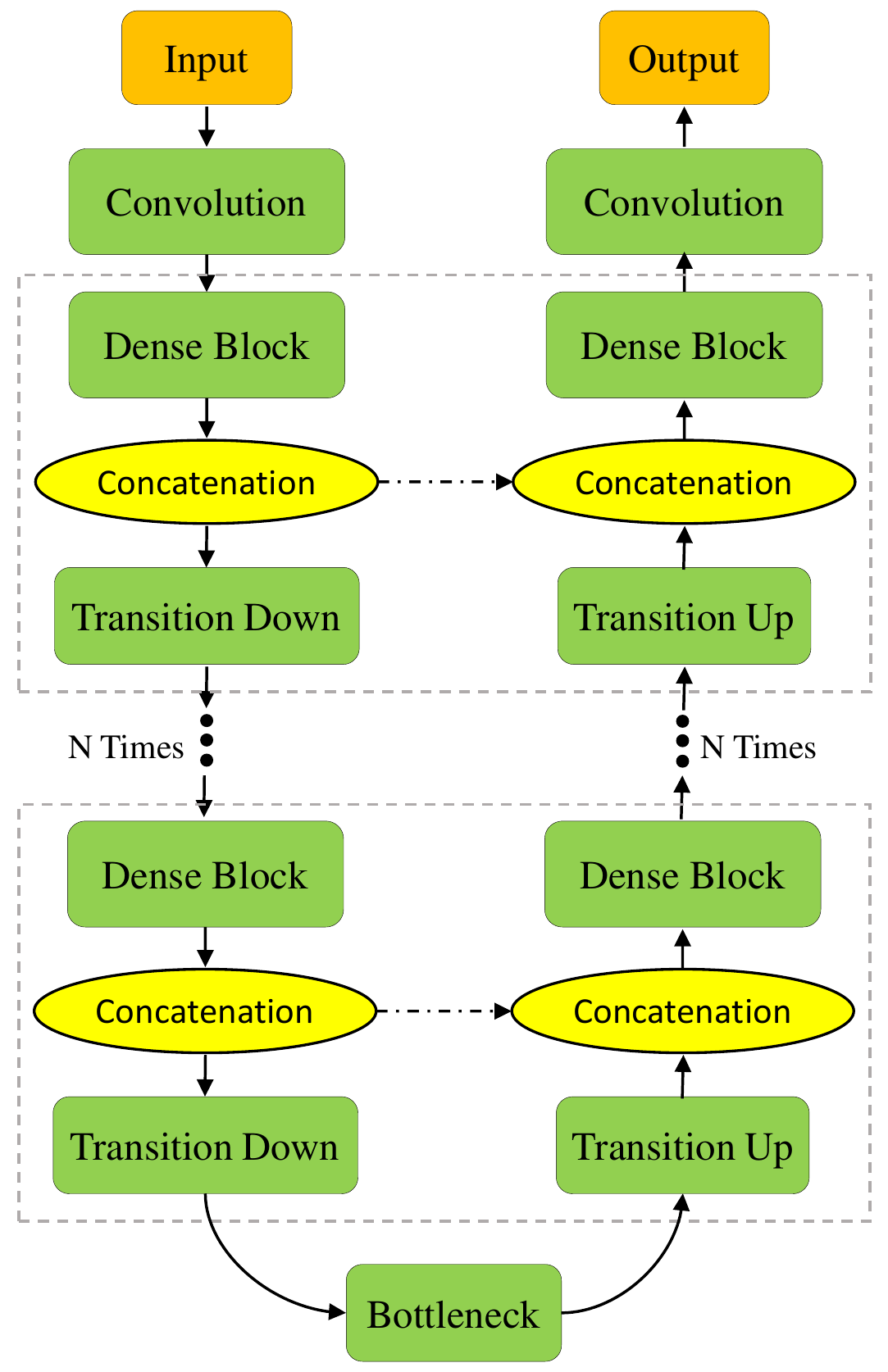}}
	%  \vspace{2.0cm}
	\caption{Diagram of main architecture of Unet with dense blocks.}
	\label{fig:FC-DenseNet}
\end{figure}

\subsection{Ensemble Modeling}
\subsubsection{Network architecture}
As shown in Fig \ref{fig:FC-DenseNet}, the main architecture comprises two major parts: contracting path and expansive path. In the contracting path, we have designed $N$ encoding stages ($N=5$ in the experiment). Each stage has three components: 1) A dense block, which is used for feature extraction; 2) A concatenation operation that gathers all feature information from the preceding dense block; 3) A transition down layer, which doubles the size of receptive fields while generating coarser feature maps. In the expansive path, corresponding decoding stages are constructed consisting of transition up layer, concatenation operation and dense block, in order to progressively recover the resolution of feature maps. Skip connections are added between corresponding stages of contracting and expansive paths, which helps to preserve contextual information for better object recognition. 
\subsubsection{Multi-path Ensemble Model}
During the experiment, we discover that the model behaves differently when adding deformable convolutional layers in contracting and expansive paths, respectively. Using model visualization tools, we find that the feature maps in contracting path are more related to contextual information and those in expansive path are more related to positional and morphological information. Therefore, as shown in Fig \ref{fig:fig_flowchart}, we train three networks in a multi-path fashion: 1) the plain network; 2) a network with deformable convolutions in contracting path; 3) a network with deformable convolutions in expansive path. We do not construct a network by replacing all normal convolutional layers with the deformable version due to the excessive computational burden.
\begin{figure*}[!ht]
	\centering
	\centerline{\includegraphics[width=0.85\linewidth]{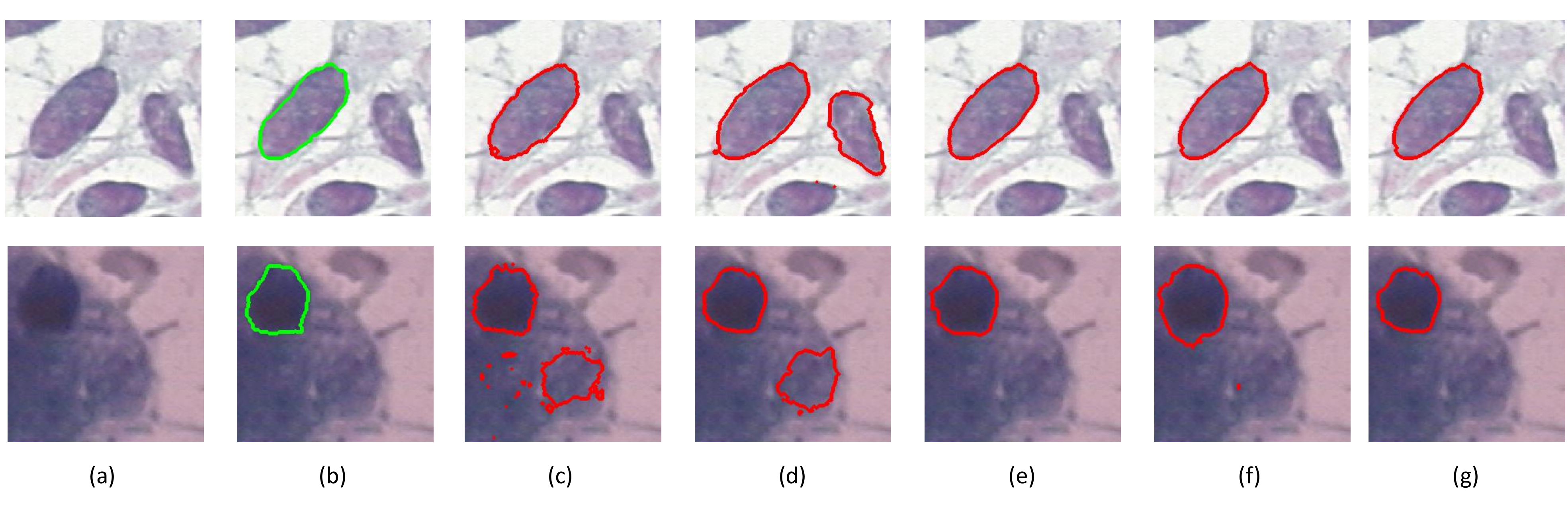}}
	%  \vspace{2.0cm}
	\caption{Examples of the segmentation results. (a) Pap smear images, (b) Manual annotations, (c) Segmentation results of Unet, (d) Segmentation results of Dense-Unet, (e) Segmentation results of D-Con, (f) Segmentation results of D-Exp, (g) Segmentation results of D-MEM.}
	\label{fig:FC-DenseNet-result}
\end{figure*}
\subsection{Model Training and Predicting}
As described in \cite{zhang2017combining}, there exists substantial difference between abnormal and normal nuclei in shape and size. We thus separate the nuclei class into abnormal (small) and normal (large) classes, resulting in four classes for D-MEM training and predicting. The proposed D-MEM is trained end-to-end and pixel-to-pixel. During the inference/prediction stage, the model produces a 4-channel prediction map, each for one of the classes (background, cervical cytoplasm, normal nuclei and abnormal nuclei). The final nuclei segmentation is obtained by combining the abnormal and normal nucleus results.

\section{Experiment and Results}
\label{sec:Experiment}
\begin{table}[!tbp]
	\centering%\makegapedcells
	\begin{tabular*}{\linewidth}{lp{1.4cm}cp{1.4cm}cp{1.4cm}cp{1.4cm}}
		%\toprule
		\hline
		Method & ZSI & Precision & Recall \\ 
		\hline
		Unsupervised\cite{gencctav2012unsupervised} & 0.89$\pm$0.15   & 0.88$\pm$0.15 & 0.93$\pm$0.15 \\%
		FCM\cite{chankong2014automatic} &0.80$\pm$0.24   & 0.85$\pm$0.21 & 0.83$\pm$0.25  \\
		P-MRF\cite{zhao2016automatic} &0.93$\pm$0.03   & $-$ & $-$  \\
		SP-CNN\cite{gautam2018cnn} &0.90   & $0.89$ & $0.91$ \\
		\textbf{Our Method} & \textbf{0.933}$\pm$0.14   & \textbf{0.946}$\pm$0.06 & \textbf{0.984}$\pm$0.00  \\%
		\hline
		%\bottomrule
	\end{tabular*}
	\caption{Comparison of the state-of-the-art methods and proposed method}
	\label{tab:table1}
\end{table}

The Herlev dataset consists of 917 cytological images from Pap tests. The original images are manually segmented into 4 labels: background, cytoplasm, nuclei and unknown regions (in this study, the unknown regions are treated as background). All images are normalized to have zero mean with unit variance intensity and are resized to a size of 256$\times$256. We train our model on a single NVIDIA GTX 1080ti GPU.

Table \ref{tab:table1} shows the quantitative comparison of the start-of-the-art methods and our proposed method in terms of mean ($\pm$ standard deviation) of ZSI, precision and recall for all images from Herlev dataset. The overall processing time of D-MEM is less than 0.1 seconds. 

We also compare the evaluation metrics computed based on different model settings. As shown in Table \ref{tab:table2}, dense block, deformable convolutional layers and ensemble modeling are able to improve the segmentation of cervical nuclei. A parallel ensemble strategy can reduce the predictive bias from training a single network Fig \ref{fig:FC-DenseNet-result} shows some examples of the segmentation results of our proposed method.
\begin{table}[!t]
	\centering%\makegapedcells
	\setlength{\tabcolsep}{1.9mm}{
	\small
	\begin{tabular*}{\linewidth}{c c c c c c}
	%\begin{tabular*}{\linewidth}{lp{0.8cm}|p{1.6cm}|p{0.9cm}|p{0.9cm}|p{1.5cm}}
		%\toprule
		\hline
		Methods & Unet & Dense-Unet & D-Con & D-Exp & \textbf{D-MEM} \\ 
		\hline
		ZSI &0.869 & 0.910  & 0.918 & 0.917 & \textbf{0.933} \\
		Precision &0.897 & 0.893  & 0.888 & 0.894 & \textbf{0.946} \\
		Recall &0.879 & 0.956  & 0.972 & 0.961 & \textbf{0.984} \\
		F-score &0.888 & 0.924 & 0.928 & 0.926 & \textbf{0.965} \\
		\hline
		%\bottomrule
	\end{tabular*}}
	\caption{Evaluation of different model settings. (D-Con, D-Exp, D-MEM stand for the models with deformable convolution on contracting path, deformable convolution on expansive path and ensemble modeling, respectively.)}
	\label{tab:table2}
\end{table}

\section{Conclusion}
\label{sec:Conclusion}
In this work, we present a novel deep-learning framework for automated segmentation of cervical nuclei. We exploit dense blocks to overcome the information vanishing problem in traditional network structures. We add deformable convolutions, so that the model could segment objects with different shapes and sizes more accurate. We also introduce ensemble modeling strategy to reduce the predictive bias from training a single network (or model setting). All these designs show improvement on the overall accuracy of segmenting cervical nuclei. The experimental results demonstrate the superior performance of the proposed method. 

\section{Acknowledgment}
\label{sec:Acknowledgment}
This work is supported in part by the National Key Research and Development Program of China under Grant 2018YFC0910700 and the National Natural Science Foundation of China (NSFC) under Grants 81801778, 11831002, 11701018. 

%\newpage
%\section{References}
%\label{sec:References}
\bibliographystyle{IEEEbib}
\bibliography{strings,refs}

\begin{thebibliography}{10}

\bibitem{insinga2004diagnoses}
Ralph~P Insinga, Andrew~G Glass, and Brenda~B Rush,
\newblock ``Diagnoses and outcomes in cervical cancer screening: a
  population-based study,''
\newblock {\em American journal of obstetrics and gynecology}, vol. 191, no. 1,
  pp. 105--113, 2004.

\bibitem{li2012cytoplasm}
Kuan Li, Zhi Lu, Wenyin Liu, and Jianping Yin,
\newblock ``Cytoplasm and nucleus segmentation in cervical smear images using
  radiating gvf snake,''
\newblock {\em Pattern Recognition}, vol. 45, no. 4, pp. 1255--1264, 2012.

\bibitem{huang2010effective}
Po-Whei Huang and Yan-Hao Lai,
\newblock ``Effective segmentation and classification for hcc biopsy images,''
\newblock {\em Pattern Recognition}, vol. 43, no. 4, pp. 1550--1563, 2010.

\bibitem{plissiti2011automated}
Marina~E Plissiti, Christophoros Nikou, and Antonia Charchanti,
\newblock ``Automated detection of cell nuclei in pap smear images using
  morphological reconstruction and clustering,''
\newblock {\em IEEE Transactions on information technology in biomedicine},
  vol. 15, no. 2, pp. 233--241, 2011.

\bibitem{bamford1998unsupervised}
Pascal Bamford and Brian Lovell,
\newblock ``Unsupervised cell nucleus segmentation with active contours,''
\newblock {\em Signal processing}, vol. 71, no. 2, pp. 203--213, 1998.

\bibitem{wang2005computerized}
Sheng-Lan Wang, Ming-Tsang Wu, Sheau-Fang Yang, Hon-Man Chan, and Chee-Yin
  Chai,
\newblock ``Computerized nuclear morphometry in thyroid follicular neoplasms,''
\newblock {\em Pathology international}, vol. 55, no. 11, pp. 703--706, 2005.

\bibitem{zhang2014segmentation}
Ling Zhang, Hui Kong, Chien~Ting Chin, Shaoxiong Liu, Zhi Chen, Tianfu Wang,
  and Siping Chen,
\newblock ``Segmentation of cytoplasm and nuclei of abnormal cells in cervical
  cytology using global and local graph cuts,''
\newblock {\em Computerized Medical Imaging and Graphics}, vol. 38, no. 5, pp.
  369--380, 2014.

\bibitem{song2014deep}
Youyi Song, Ling Zhang, Siping Chen, Dong Ni, Baopu Li, Yongjing Zhou, Baiying
  Lei, and Tianfu Wang,
\newblock ``A deep learning based framework for accurate segmentation of
  cervical cytoplasm and nuclei,''
\newblock in {\em Engineering in Medicine and Biology Society (EMBC), 2014 36th
  annual international conference of the IEEE}. IEEE, 2014, pp. 2903--2906.

\bibitem{song2015accurate}
Youyi Song, Ling Zhang, Siping Chen, Dong Ni, Baiying Lei, and Tianfu Wang,
\newblock ``Accurate segmentation of cervical cytoplasm and nuclei based on
  multiscale convolutional network and graph partitioning,''
\newblock {\em IEEE Transactions on Biomedical Engineering}, vol. 62, no. 10,
  pp. 2421--2433, 2015.

\bibitem{sharma2016improved}
Bharti Sharma and Kamaljeet~Kaur Mangat,
\newblock ``An improved nucleus segmentation for cervical cell images using fcm
  clustering and bpnn,''
\newblock in {\em Advances in Computing, Communications and Informatics
  (ICACCI), 2016 International Conference on}. IEEE, 2016, pp. 1924--1929.

\bibitem{zhang2017combining}
Ling Zhang, Milan Sonka, Le~Lu, Ronald~M Summers, and Jianhua Yao,
\newblock ``Combining fully convolutional networks and graph-based approach for
  automated segmentation of cervical cell nuclei,''
\newblock in {\em Biomedical Imaging (ISBI 2017), 2017 IEEE 14th International
  Symposium on}. IEEE, 2017, pp. 406--409.

\bibitem{jantzen2005pap}
Jan Jantzen, Jonas Norup, Georgios Dounias, and Beth Bjerregaard,
\newblock ``Pap-smear benchmark data for pattern classification,''
\newblock {\em Nature inspired Smart Information Systems (NiSIS 2005)}, pp.
  1--9, 2005.

\bibitem{jegou2017one}
Simon J{\'e}gou, Michal Drozdzal, David Vazquez, Adriana Romero, and Yoshua
  Bengio,
\newblock ``The one hundred layers tiramisu: Fully convolutional densenets for
  semantic segmentation,''
\newblock in {\em Computer Vision and Pattern Recognition Workshops (CVPRW),
  2017 IEEE Conference on}. IEEE, 2017, pp. 1175--1183.

\bibitem{ouyang2015deepid}
Wanli Ouyang, Xiaogang Wang, Xingyu Zeng, Shi Qiu, Ping Luo, Yonglong Tian,
  Hongsheng Li, Shuo Yang, Zhe Wang, Chen-Change Loy, et~al.,
\newblock ``Deepid-net: Deformable deep convolutional neural networks for
  object detection,''
\newblock in {\em Proceedings of the IEEE Conference on Computer Vision and
  Pattern Recognition}, 2015, pp. 2403--2412.

\bibitem{gencctav2012unsupervised}
Asl{\i} Gen{\c{c}}Tav, Selim Aksoy, and Sevgen {\"O}Nder,
\newblock ``Unsupervised segmentation and classification of cervical cell
  images,''
\newblock {\em Pattern recognition}, vol. 45, no. 12, pp. 4151--4168, 2012.

\bibitem{chankong2014automatic}
Thanatip Chankong, Nipon Theera-Umpon, and Sansanee Auephanwiriyakul,
\newblock ``Automatic cervical cell segmentation and classification in pap
  smears,''
\newblock {\em Computer methods and programs in biomedicine}, vol. 113, no. 2,
  pp. 539--556, 2014.

\bibitem{zhao2016automatic}
Lili Zhao, Kuan Li, Mao Wang, Jianping Yin, En~Zhu, Chengkun Wu, Siqi Wang, and
  Chengzhang Zhu,
\newblock ``Automatic cytoplasm and nuclei segmentation for color cervical
  smear image using an efficient gap-search mrf,''
\newblock {\em Computers in biology and medicine}, vol. 71, pp. 46--56, 2016.

\bibitem{gautam2018cnn}
Srishti Gautam, Arnav Bhavsar, Anil~K Sao, and KK~Harinarayan,
\newblock ``Cnn based segmentation of nuclei in pap-smear images with selective
  pre-processing,''
\newblock in {\em Medical Imaging 2018: Digital Pathology}. International
  Society for Optics and Photonics, 2018, vol. 10581, p. 105810X.

\end{thebibliography}

\end{document}